\documentclass[journal]{IEEEtran}
\usepackage{amsmath,amsfonts}
\usepackage{algorithmic}
\usepackage{algorithm}
\usepackage{array}
\usepackage[caption=false,font=normalsize,labelfont=sf,textfont=sf]{subfig}
\usepackage{textcomp}
\usepackage{stfloats}
\usepackage{url}
\usepackage{verbatim}
\usepackage{graphicx}
\usepackage{cite}
\usepackage{hyperref}
\usepackage{gensymb}
\usepackage[percent]{overpic}
\usepackage{multirow}
\usepackage{array}
\usepackage{amssymb}
\usepackage{caption}
\usepackage{tikz}
\usepackage{pdfpages}
\newcommand\copyrighttext{%
  \footnotesize 
This work is currently under review in a IEEE journal
}
\newcommand\copyrightnotice{%
\begin{tikzpicture}[remember picture,overlay]
\node[anchor=south,yshift=10pt] at (current page.south) {\fbox{\parbox{\dimexpr\textwidth-\fboxsep-\fboxrule\relax}{\copyrighttext}}};
\end{tikzpicture}%
}

\begin{document}

\title{ \textit{ros2\_fanuc\_interface}: Design and Evaluation of a Fanuc CRX Hardware Interface in ROS2}

\author{Paolo Franceschi$^1$, Marco Faroni$^2$, Stefano Baraldo$^1$, Anna Valente$^1$
\thanks{Corresponding author: {\tt\small paolo.franceschi@supsi.ch }}
\thanks{$^1$ Department of Innovative Technologies, University of Applied Science and Arts of Southern Switzerland (SUPSI), Lugano, Switzerland}
\thanks{$^2$ Dipartimento di Elettronica, Informazione e Bioingegneria, Politecnico di Milano, Milan, Italy}
}



\maketitle
\copyrightnotice

\begin{abstract}
This paper introduces the ROS2 control and hardware interface integration for the Fanuc CRX robot family.
It explains basic implementation details and communication protocols, and its integration with the Moveit2 motion planning library.
We conducted a series of experiments to evaluate relevant performances in the robotics field.
We tested the developed \textit{ros2\_fanuc\_interface} for four relevant robotics cases: step response, trajectory tracking, collision avoidance integrated with Moveit2, and dynamic velocity scaling, respectively.
Results show that, despite a non-negligible delay between command and feedback, the robot can track the defined path with negligible errors (if it complies with joint velocity limits), ensuring collision avoidance.
Full code is open source and available at \href{https://github.com/paolofrance/ros2_fanuc_interface}{https://github.com/paolofrance/ros2\_fanuc\_interface}.

\end{abstract}


\section{Introduction}
\IEEEPARstart{T}{he} Robot Operating System (ROS) ecosystem is the de facto standard to control robots in the research field and is growing in interest from the industrial world.
ROS and ROS2 allow for seamless integration of state-of-the-art motion planning, vision, and control algorithms, acting as a middleware for modular software development and deployment.
While initially widespread mainly in academia, ROS2 has now reached a level of maturity that makes it usable in industrial applications \cite{bonci2023robot}.

In the ROS2 world, the \textit{ros2\_control} framework \cite{Chitta2017} is a control architecture designed to standardize and simplify the development of controllers and hardware abstractions.
\textit{ros2\_control} separates high-level control logic from low-level hardware interfaces, promoting reusability and maintainability across diverse robotic systems.
Central to this architecture are the \textit{ros2\_controllers} package and the concept of \textit{hardware interfaces}.
On the one hand, \textit{ros2\_controllers} provides a collection of standardized controllers for common robotic functionalities (such as joint trajectory tracking, forward command execution, and state broadcasting, etc.) designed to work seamlessly with different robot configurations through well-defined interfaces.
On the other hand, \textit{hardware interface} defines how actuator and sensor data are exchanged between ROS2 and the physical robot. 
In other words, it translates the standard interfaces expected by ROS2 controllers into hardware-dependent commands and data formats of a specific robot.

Most industrial robots are integrated, to some extent, with the \textit{ros2\_control} framework \cite{doi:10.1126/scirobotics.abm6074}. 
For example, commonly used robots, such as KUKA LBR \cite{Huber2024, IIWAros2}, Universal Robots \cite{UR}, ABB \cite{ABB}, Franka \cite{franka}, have their \textit{hardware interfaces} freely available.
Unfortunately, this is not the case for the Fanuc CRX Collaborative Robot Series.

To the best of the authors' knowledge, \cite{fanucros2driver} is the only attempt to integrate Fanuc robots with ROS2. 
Such a package focuses on developing Python APIs integrated with ROS2 high-level actions. However, its Python implementation is not compatible with \textit{ros2\_control} and hinders real-time performance. 

\begin{figure}[t]
    \centering
    \begin{overpic}[width=0.9\columnwidth]{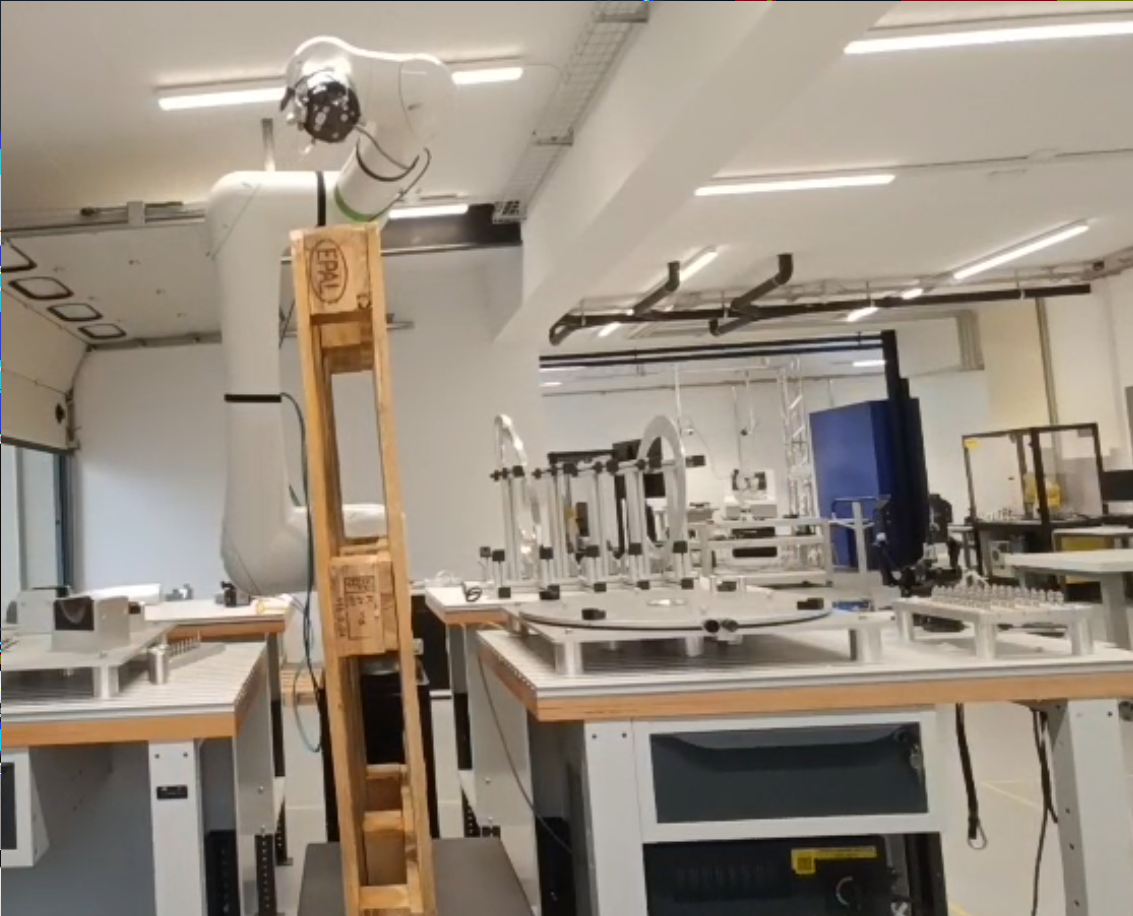}
        \put(50,0){\includegraphics[width=0.45\columnwidth]{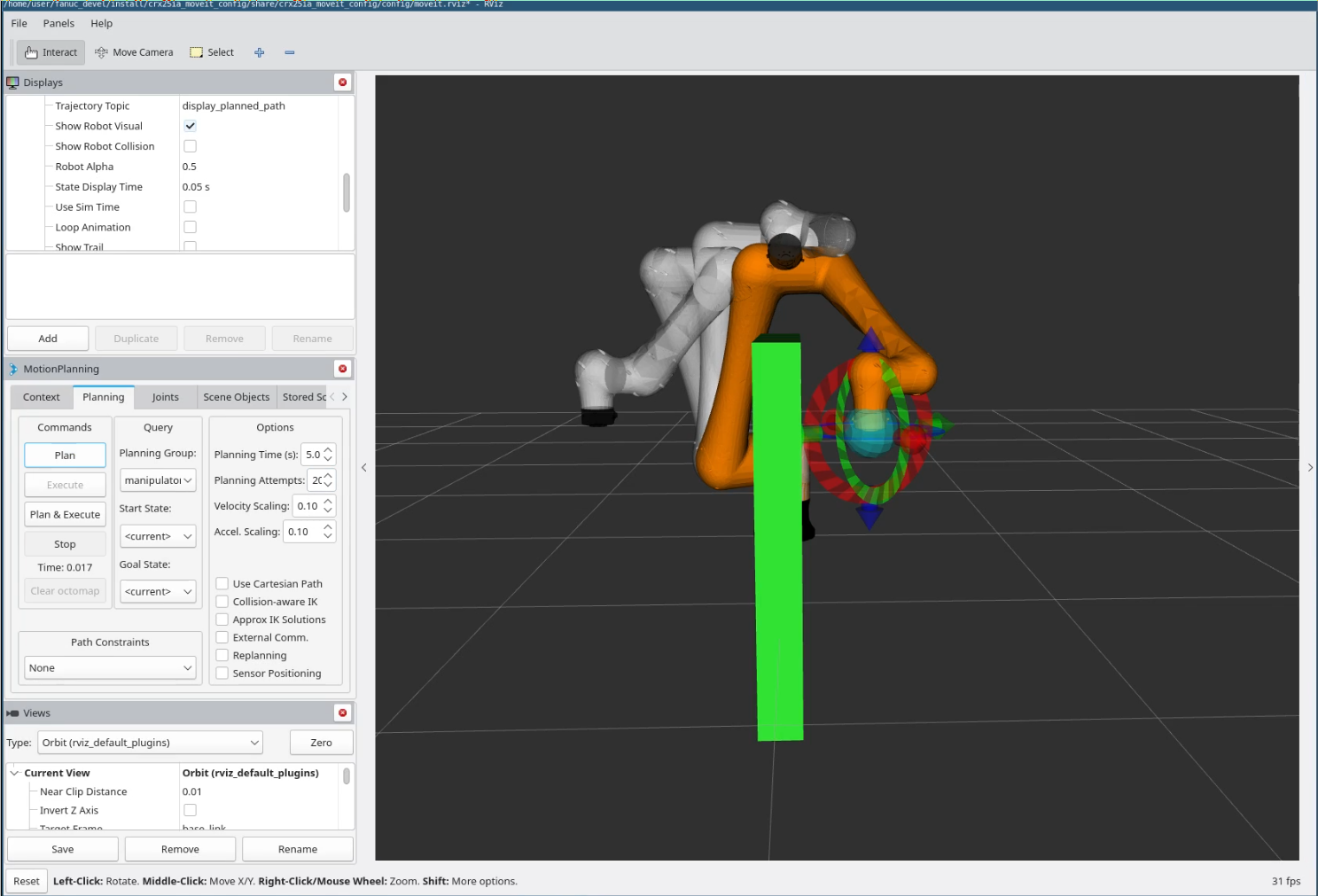}}  
    \end{overpic}
    \caption{Real and simulated scene of the collision avoidance experiment.}
    \label{fig:moveit_scene}
\end{figure}

This work presents and validates the usage of the Fanuc CRX family within the \textit{ros2\_control} framework. 
It describes the \textit{hw\_interface} implementation based on the Ethernet IP communication protocol and its integration with ROS2 tools such as MoveIt2 \cite{moveit} (see Figure \ref{fig:moveit_scene}).
It also experimentally evaluates the control performance and discusses the limitations of the proposed approach.
The additional material describes the open-source software packages available at \href{https://github.com/paolofrance/ros2_fanuc_interface}{https://github.com/paolofrance/ros2\_fanuc\_interface}.

\section{Preliminaries}

When developing a \textit{hardware interface} from scratch, it is essential to understand how a remote computer communicates with the robot controller.
%
%
The Fanuc CRX robots are managed by the Fanuc R30iB Mini Plus controller, which allows for remote commands to the robot through two modalities, Remote Motion Interface (RMI) and Ethernet/IP (EIP). 

RMI allows control of the robot's motion by sending commands from a remote PC to the controller in JSON format. 
A special teach pendant (TP) program appends and executes the instructions received from the remote PC to the TP program as soon as they are available.
The TP program executes the commands sequentially, removing the ones already executed to let new ones incoming, with a buffer of eight slots.
RMI communicates through TCP/IP socket messages.

The EIP interface supports I/O exchange with the remote computer over an Ethernet network.
EIP allows easy access to several Fanuc signals, including reading and writing numeric Registers (R), Position Registers (PR), and Digital Inputs (DI).

RMI represents a powerful solution to use Fanuc functions remotely, but it is not meant for streaming position commands, and has a limited number of instructions available.
Therefore, our hardware interface relies on the EIP interface.

\section{ROS2-Fanuc Hardware Interface}

This section describes the main components of the ROS2 Fanuc interface; namely, how we use the Fanuc EIP interface to communicate with the robot and how the robot executes motion commands received from the external computer. 

\subsection{Fanuc Ethernet/IP}

The Fanuc Ethernet/IP communication module provides a direct interface with the Fanuc R-30iB Mini Plus controller, enabling remote access to internal memory structures such as Registers (R) and Position Registers (PR). 
Position Registers serve as memory slots that can store either joint-space or Cartesian-space target poses.

In the proposed control architecture, a ROS2-based driver running on an external computer communicates with the robot by writing desired target poses into PR[1], at a frequency specified by the \textit{ros2\_control} hardware interface.
The motion execution logic is implemented in a TP program on the Fanuc controller according to the value stored in PR[1].

The Ethernet/IP channel also continuously communicates the current robot position to the external computer and the states of digital signals.

The Etherner/IP module also allows calling additional functionalities, such as controlling a tool-mounted gripper, that can be implemented using mapped registers.

Examples of available ROS2 services implementing such features can be found in the \textit{fanuc\_srvs} package described in the additional material attached to this paper.

\subsection{TP program}

\begin{figure}[tpb]
\vspace{-0.5cm}
\centering
\includegraphics[width=\columnwidth]{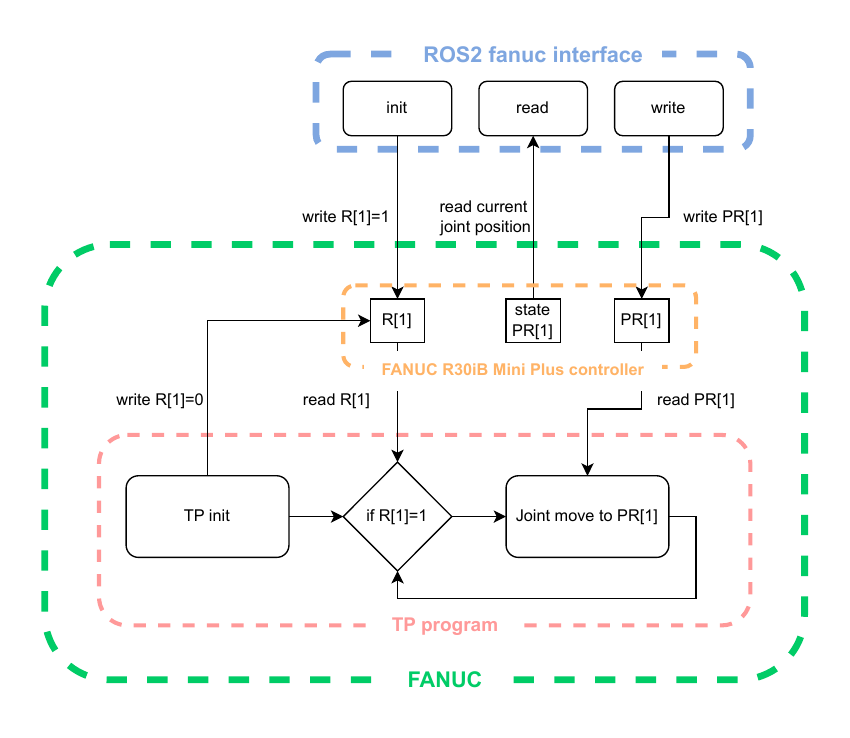}
\vspace{-1cm}
\caption{Scheme of the interaction between the TP program running on the robot controller and ROS2 hardware interface developed in \textit{ros2\_fanuc\_interface}.}
\vspace{-0.2cm}
\label{fig:TP_fanuc}
\end{figure}

The Hardware Interface requires a TP program running on the Fanuc controller to move the robot.
The logic behind the basic TP program and hardware interface implementation is in Figure \ref{fig:TP_fanuc}.

The combination of the TP program and the ROS2 hardware interface works as follows:
\begin{enumerate}
    \item When started, the TP program resets Register R[1].
    \item The TP program waits for R[1] to be set by the external computer, as a means to wait for the ROS2 hardware interface to be ready. This mechanism provides a straightforward interface for coordinating motion triggers from the external control system, ensuring that the hardware interface initializes with the current robot pose before sending commands.
    \item The hardware interface sets R[1]=1 at the end of the initialization procedure. Then, it continuously writes position commands to PR[1].
    \item If R[1]==1, the TP program continuously commands the robot to move to the target joint position available in PR[1]. Using a CNT=100\% motion instruction enables smooth trajectory execution.
\end{enumerate}

This register-based if-then-else structure of the TP program can be extended to use other registers to command other non-motion operations.
For example, one may use R[2] to command the opening/closing of a gripper through a digital input signal.
Several services, such as the gripper ones in the \textit{fanuc\_srvs}, work according to this principle.

\section{Performance evaluations}

To check the performance and usability of the developed hardware interface, we conducted experiments for four relevant tasks in industrial robotics: reaching a target point, tracking a trajectory, executing a collision-free trajectory, and dynamic velocity scaling.
Experiments are performed with a Fanuc CRX-25iA communicating via Ethernet with a remote PC at a frequency of $25Hz$.
Position and velocity measurements were recorded at the ROS2 controller's level, which means they include communication latency and ROS2's possible computational overheads.
Comparable results were observed with a CRX-10iA\_L and CRX-20iA\_L robots.

\subsection{Step response}

We measured the position step response of the robot controller.
For this experiment, we requested the robot to move only the first joint $J1$ from a fixed nominal position, to three different setpoints, $sp=30\degree,45\degree,90\degree$.

\begin{figure*}[tpb]
  \centering
  \subfloat[Step response, $ref=30\degree$\label{fig:step30}]{\includegraphics[width=0.30\textwidth]{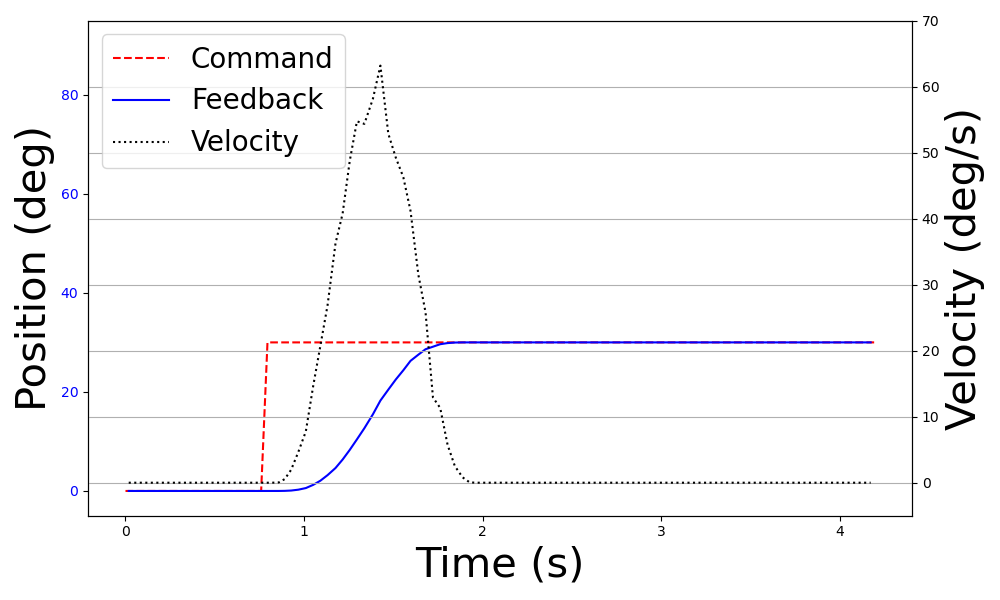}} \hfill
  \subfloat[Step response, $ref=45\degree$\label{fig:step45}]{\includegraphics[width=0.30\textwidth]{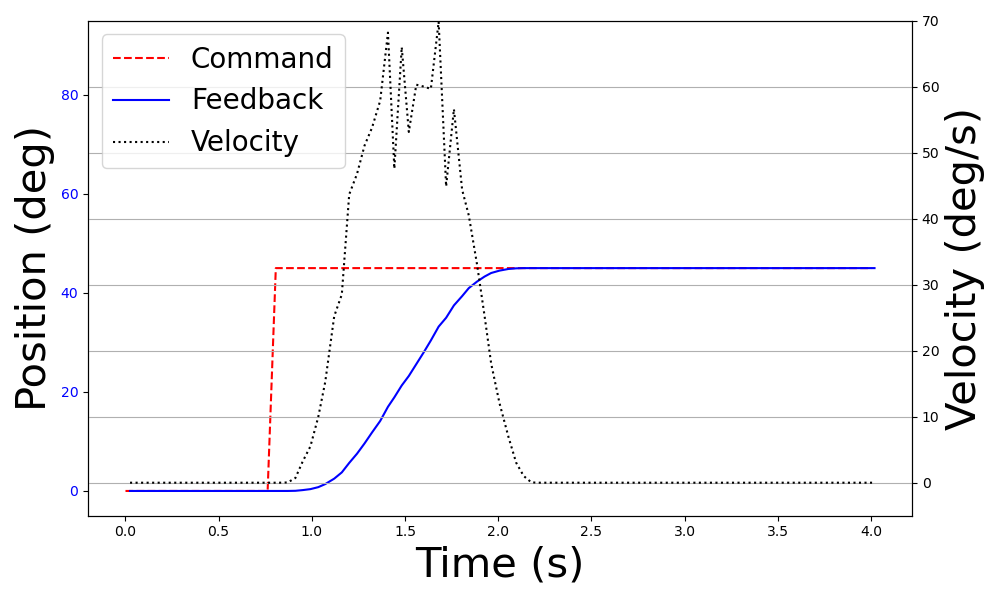}} \hfill
  \subfloat[Step response, $ref=90\degree$\label{fig:step90}]{\includegraphics[width=0.30\textwidth]{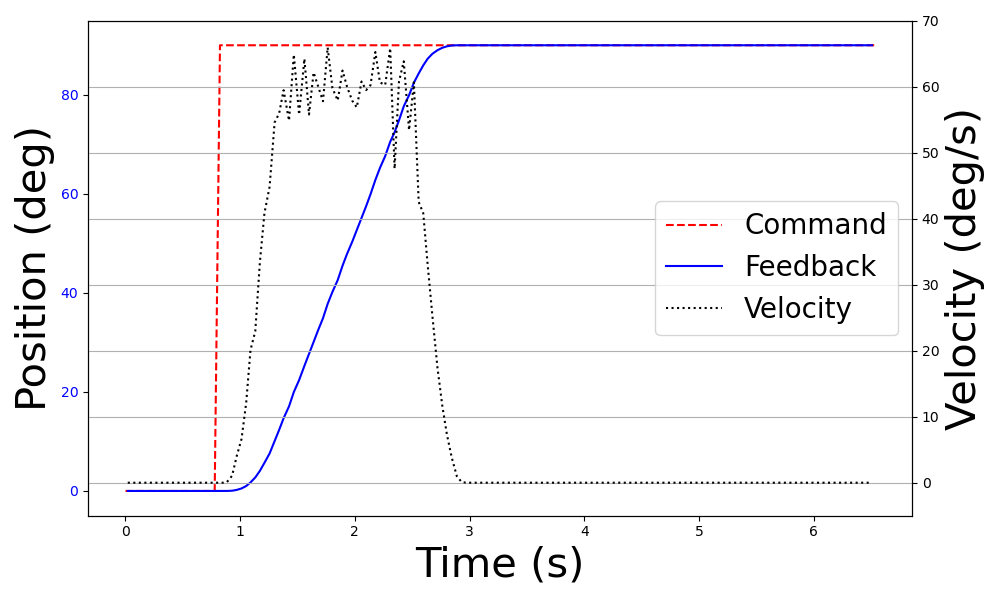}} \\
  \subfloat[Tracking response, $f=0.1Hz$\label{fig:track.1}]{\includegraphics[width=0.30\textwidth]{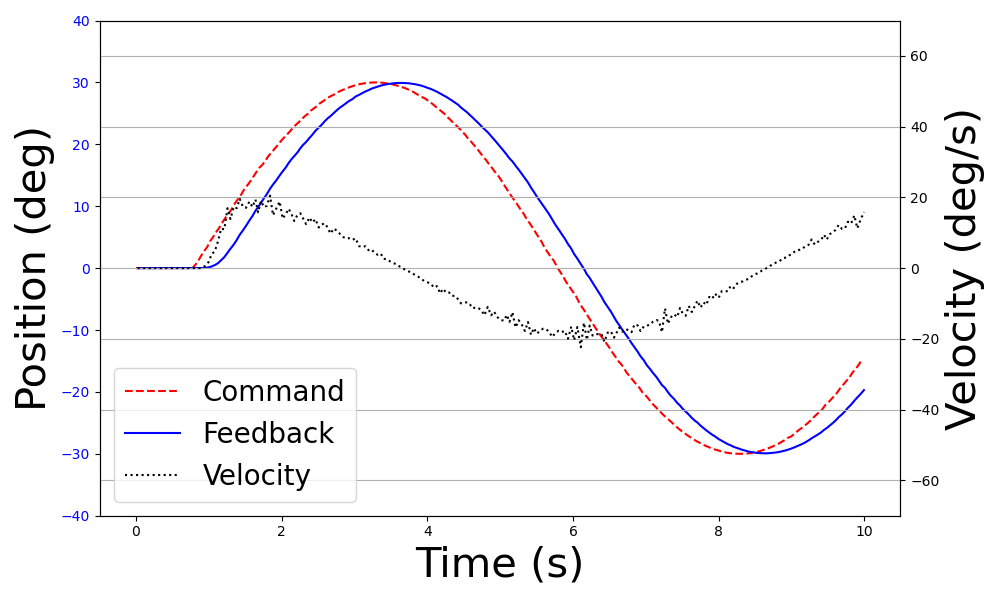}} \hfill
  \subfloat[Tracking response, $f=0.25Hz$\label{fig:track.25}]{\includegraphics[width=0.30\textwidth]{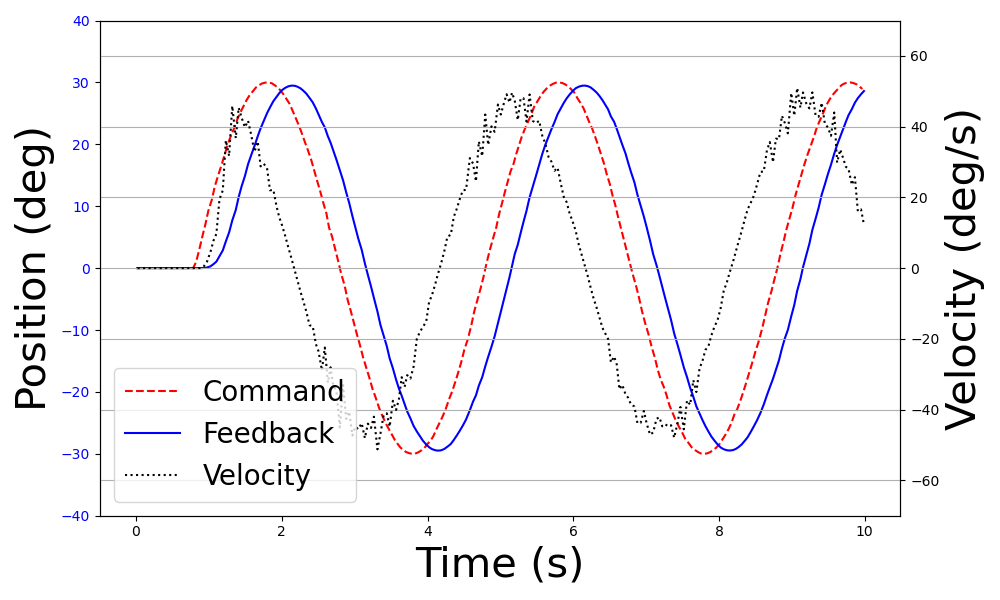}} \hfill
  \subfloat[Tracking response, $f=0.5Hz$\label{fig:track.5}]{\includegraphics[width=0.30\textwidth]{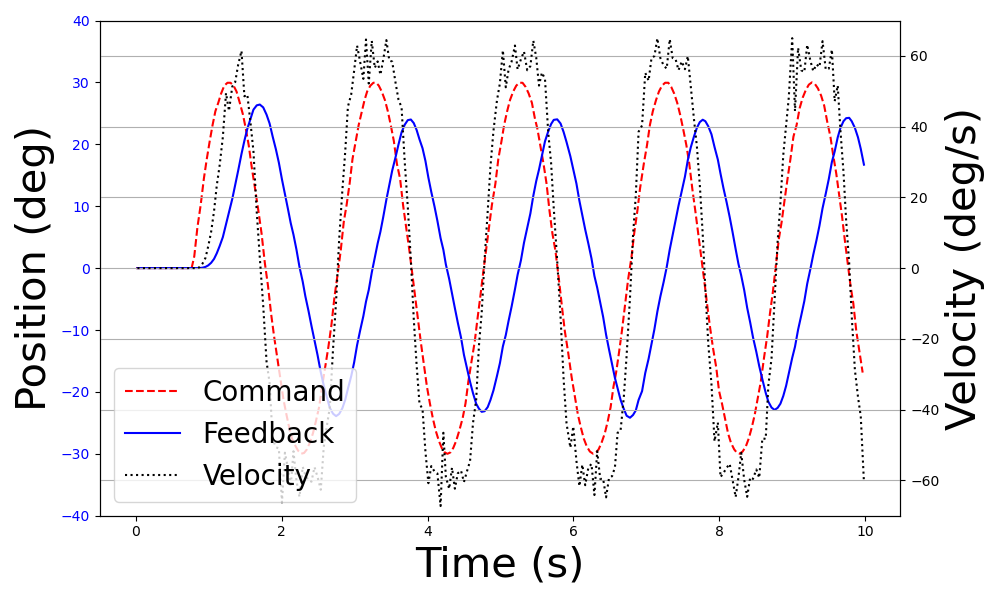}} \\
  \subfloat[Aligned tracking response, $f=0.1Hz$\label{fig:path.1}]{\includegraphics[width=0.30\textwidth]{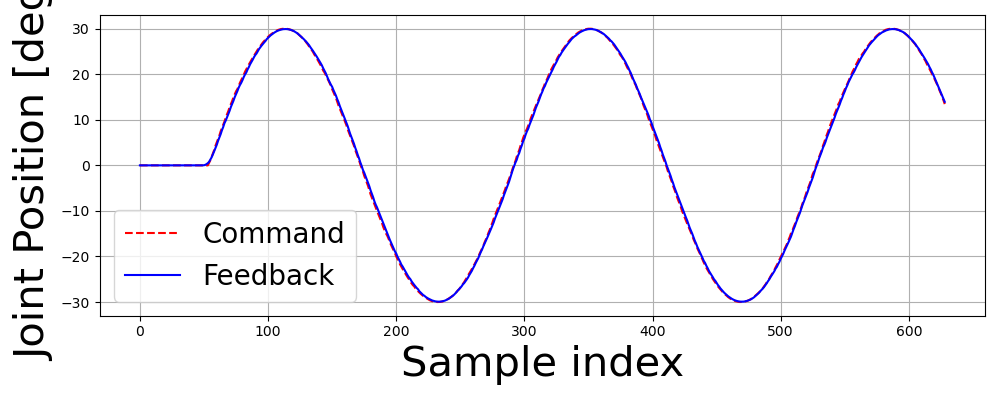}} \hfill
  \subfloat[Aligned tracking response, $f=0.25Hz$\label{fig:path.25}]{\includegraphics[width=0.30\textwidth]{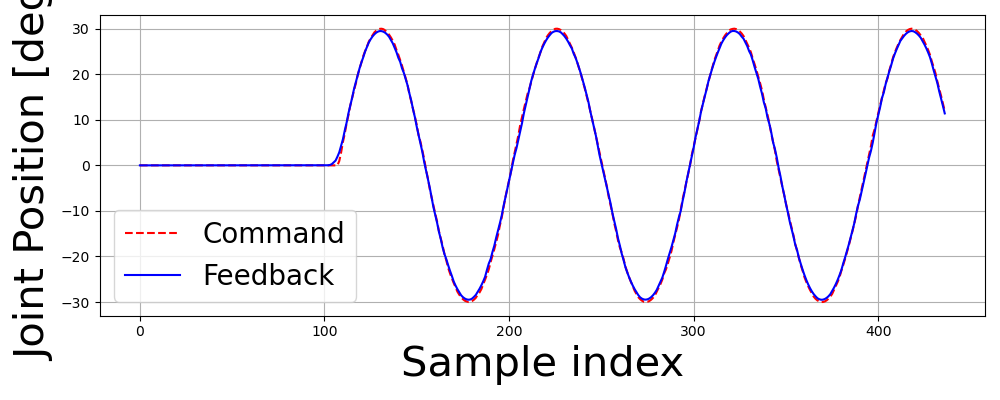}} \hfill
  \subfloat[Aligned tracking response, $f=0.5Hz$\label{fig:path.5}]{\includegraphics[width=0.30\textwidth]{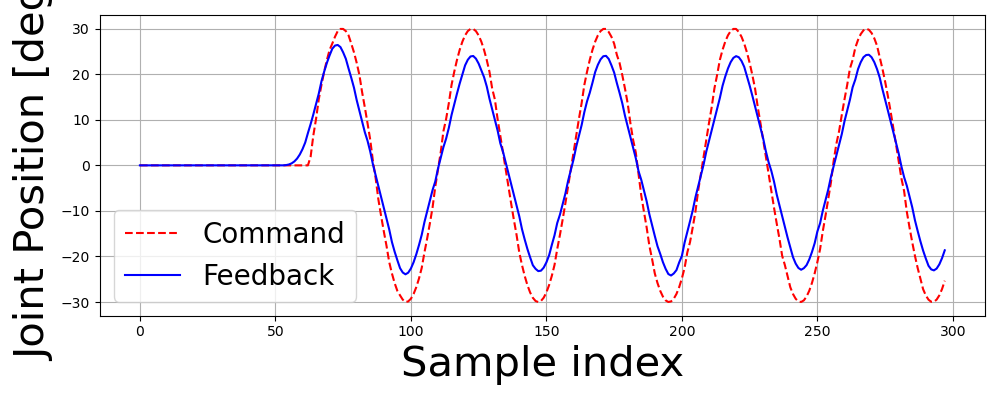}}
  \caption{Results of the step responses (top), trajectory tracking (middle), and path following (bottom).}
  \label{fig:experiments}
  \vspace{-0.3cm}
\end{figure*}

We evaluate standard performance indicators such as rise time $t_r$, settling time $T_s$, overshoot percentage $\%OS$, steady state error $err_{ss}$, and peak time $t_p$.
Results are in Table \ref{tab:step_response} and Figures \ref{fig:step30}, \ref{fig:step45}, and \ref{fig:step90}.

As expected, $t_r$ and $T_s$ depend on the step amplitude.
Moreover, $T_s$ is also influenced by the joint velocity limits enforced by the Fanuc controller. 
This is visible in Figures \ref{fig:step45} and \ref{fig:step90}, where the measured velocity (dotted black line) saturates at about $60\degree$, increasing the rise time.
Interestingly, the robot does not present any overshoots and shows negligible steady-state error.
Therefore, $t_p$ coincides with the time required to reach the set point.

\begin{table}[tpb]
\caption{Step responses (in degrees).}
\centering
\begin{tabular}{|l|c|c|c|c|c|}
\hline
\textbf{Set Point [$\degree$]} & \textbf{$t_r$} & \textbf{$T_s$} & \textbf{$\%OS$}& \textbf{$err_{ss}$} & \textbf{$t_p$} \\
\hline
30 & 0.51 & 0.90 & 0.00 & 0.00 & 1.15 \\
45 & 0.64 & 1.10 & 0.00 & 0.00 & 1.35 \\
90 & 1.22 & 1.79 & 0.00 & 0.00 & 2.12 \\
\hline
\end{tabular}
\label{tab:step_response}
\end{table}

\subsection{Trajectory tracking}

To measure the tracking capabilities of the hardware interface, we streamed sinusoidal reference trajectories, $ref(t) = A\sin (2\pi f t)$, with amplitude $ A=30\degree$ and frequencies $f\in\{0.1, 0.25, 0.5\} \, Hz$. Also, for this experiment, we requested the robot to move only the first joint. We evaluate the tracking error and the path following error.
%
%

\subsubsection{Trajectory tracking error}

For each frequency, we measured the Mean Absolute Error $\text{MAE} = \frac{1}{n} \sum_{i=1}^{n} \left| q_{\mathrm{fb},i} - q_{\mathrm{cmd},i} \right|$, the Root Mean Squared Error $\text{RMSE} = \sqrt{ \frac{1}{n} \sum_{i=1}^{n} \left( q_{\mathrm{fb},i} - q_{\mathrm{cmd},i} \right)^2 }$, and the Maximum Error $ = \max_{i \in \{1, \dots, n\}} \left| q_{\mathrm{fb},i} - q_{\mathrm{cmd},i} \right|$, where $q_{\mathrm{cmd}}$ and $q_{\mathrm{fb}}$ are the position command and feedback signals for the selected joint, and $n$ is the number of measurements.

The results are in Table \ref{tab:tracking_errors} and Figures \ref{fig:track.1}, \ref{fig:track.25}, and \ref{fig:track.5}

The upper rows of Table \ref{tab:tracking_errors} highlight that the average and maximum errors increase with frequency. 
In particular, in the case of high frequency, this is due to the joint velocity limits intrinsic in the Fanuc robots. 
This is visible in Figure \ref{fig:track.5}, where the velocity profile (dotted line) saturates around $\pm60\degree$.
%

\subsubsection{Path following error}

Tracking errors are mainly caused by the communication latency between the controller command and the robot movement. 
It is meaningful to measure path following performance, i.e., the control error between the command and feedback signals after time synchronization.

The sinchronized signals are obtained by computing the cross-correlation, $R_{q_{\mathrm{cmd}},q_{\mathrm{fb}}}[\tau] = \sum_{i=0}^{n-1} q_{\mathrm{cmd}}[i] \cdot q_{\mathrm{fb}}[i + \tau]$, between the command and feedback signals. 
We then find the lag $\tau^*$ that maximizes cross-correlation between the two signals.
We apply the same metrics of the trajectory tracking case to the error signal given by
$q_{\mathrm{cmd}}[i]$ and $q_{\mathrm{fb}}[i + \tau^*]$.

In this case, for the low-frequency cases ($f=\{0.1,0.25\}Hz$), the errors dramatically decrease, with averages below $1\degree$.
As for the higher frequency, $f=0.5Hz$, the velocity of the joint saturates, and the resulting error remains large, leading to poor path following.
The maximum error is measured when the command changes from zero to $ref(t)$. 
In this case, the robot should have a non-zero velocity and must adhere to physical constraints.
The path errors are in the bottom rows of Table \ref{tab:tracking_errors} and in Figure \ref{fig:path.1}, \ref{fig:path.25}, and \ref{fig:path.5}.


\begin{table}[tpb]
\caption{Path and trajectory tracking error results.}
\centering
\begin{tabular}{|m{4em}|c|c|c|c|}
\hline
& \textbf{\textit{f} [Hz]} & \textbf{MAE} & \textbf{RMSE} & \textbf{Max Error} \\
\hline
\multirow{3}{*}{\shortstack{trajectory\\tracking}}
& 0.1 & 3.72 & 4.34 & 6.67 \\
& 0.25 & 7.77 & 9.96 & 17.03 \\
& 0.5 & 18.26 & 22.57 & 36.70 \\

\hline
\multirow{3}{*}{\shortstack{path\\following}}
& 0.1 & 0.36 & 0.42 & 0.87 \\
& 0.25 & 0.41 & 0.49 & 1.91 \\
& 0.5 & 5.22 & 5.68 & 8.78 \\

\hline
\end{tabular}
\vspace{-0.3cm}
\label{tab:tracking_errors}
\end{table}

\subsubsection{Time delay}

We measured the time delay between the command and feedback, as well as the allowed control frequency. 
Relying on the sinusoidal commands, we computed the cross-correlation to obtain the lag, for the first two cases ($f=\{0.1, 0.25\} Hz$).
Then we measured the average cycle time to obtain the time delay and control frequency.
The time delay varies a little between the cases, with an average value of about $0.31s$, and a control frequency of about $24Hz$, which is in line with Fanuc's EIP specifications.

\subsection{Obstacle avoidance}

\begin{figure}[tpb]
\includegraphics[width=\columnwidth]{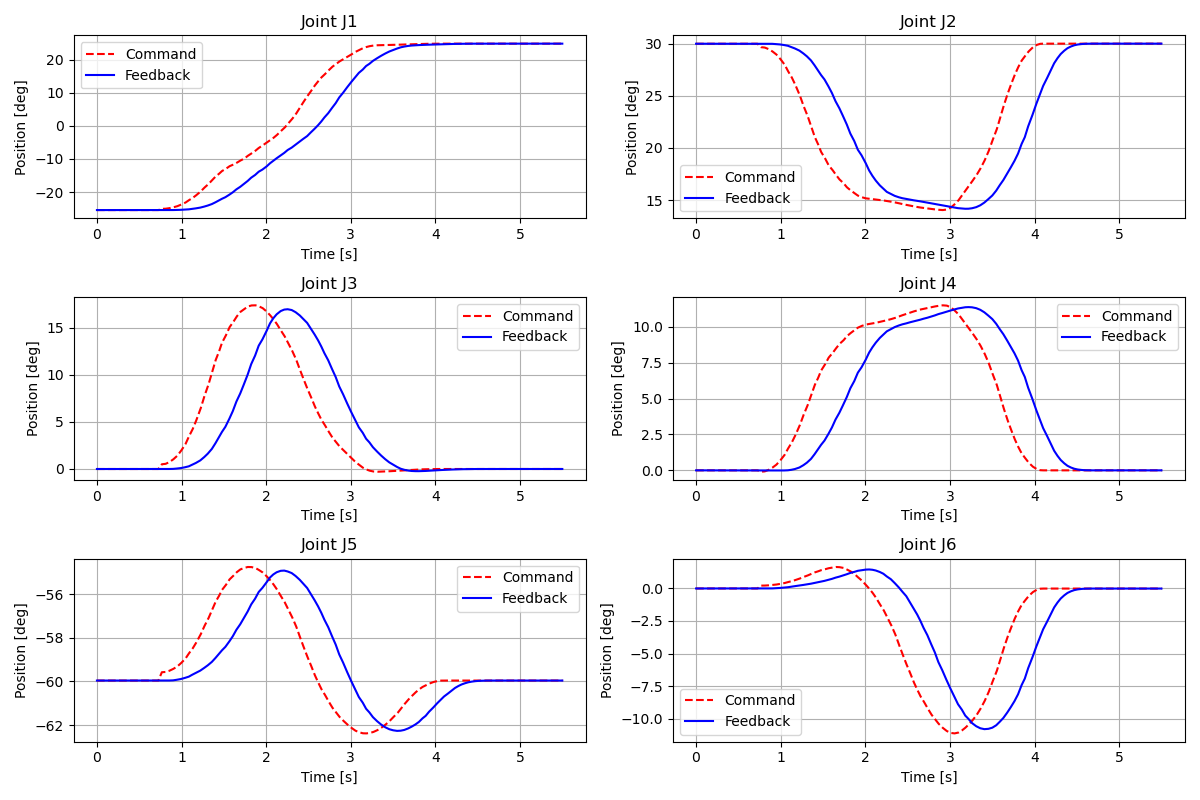}
\centering
\caption{Command and feedback signals during execution of Moveit2 computed trajectory with obstacle avoidance.}
\label{fig:moveit_joints}
\vspace{-0.3cm}
\end{figure}

We assessed the correct execution of collision-free trajectories via Moveit2 \cite{moveit} through the experiment shown in Figure \ref{fig:moveit_scene}. 
The robot has to move from an initial  
to a goal configuration by avoiding a box-shaped obstacle.
The trajectory is computed with the BiTRRT motion planner, with a velocity scaling of 0.1.
The executed trajectory is in Figure \ref{fig:moveit_joints}.
The plots confirm the results obtained in the previous experiments in terms of trajectory tracking and path following.

\subsection{Dynamic velocity scaling}

\begin{figure}[tpb]
\includegraphics[width=\columnwidth,trim={0cm 0cm 0cm 0cm},clip]{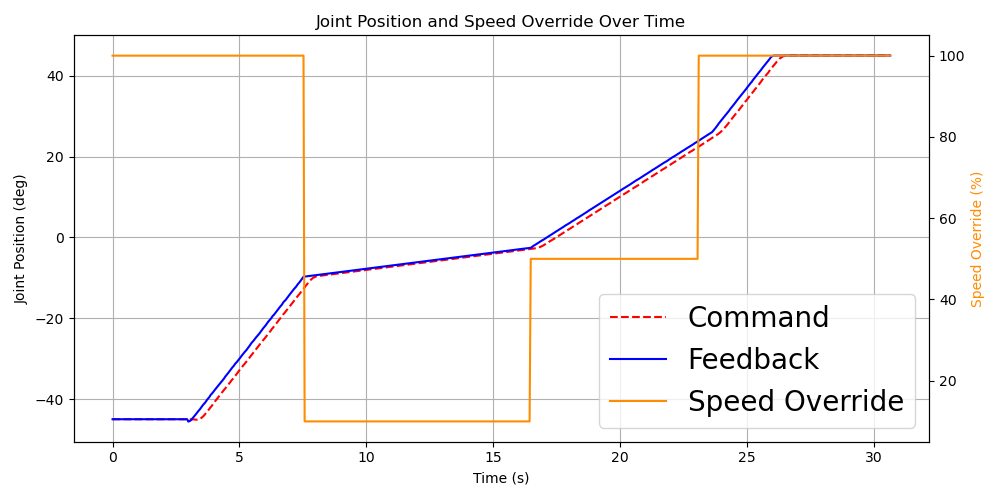}
\centering
\caption{Command and feedback signals with dynamic velocity scaling during trajectory execution. The speed override is set dynamically via ros topic.}
\label{fig:speed_ovr}
\end{figure}

We used the proposed hardware interface to implement online velocity scaling during trajectory execution.
In the proposed experiment, a trajectory is computed with Moveit2 to move the first joint from $-45\degree$ to $45\degree$, maintaining the other joints fixed.
The speed override is manually set to $10\%,\;50\%,\;100\%$. Figure \ref{fig:speed_ovr} shows the command and feedback signals, as well as the override values.
This experiment used a custom ROS2 controller, named \textit{scaled\_fjt\_controller}, based on \cite{8477138,8869047} and available here \cite{scaledfjt}.

We used this feature for safe human-robot interaction and human-robot collaborative transport applications.
Detailed experiments of human-robot collaborative transport can be found in \cite{franceschi2025humanrobotcollaborativetransportpersonalization}, where the proposed \textit{ros2\_fanuc\_interface} is used in combination with the \textit{scaled\_fjt\_controller} to adapt trajectory execution based on interaction forces.


\subsection{Limitations}
Closed-loop control is limited by the latency.
We attempted to implement force and admittance control, but achieved unsatisfactory results, as the delay led to unstable behaviors.
To implement feedback control on an external computer, we performed preliminary tests exploiting the Dynamic Path Modification (DPM) function of the Fanuc robots, but its integration in the \textit{ros\_control} is not straightforward and left as a future work.

The \textit{ros2\_fanuc\_interface} does not enforce joint velocity limits. 
This means that large velocity commands may be clamped by the Fanuc controller. 
For this reason, careful attention must be paid to the robot controller's settings, as the Fanuc controller may enforce Cartesian speed limits of around 250 mm/s in the collaborative mode.
This may lead to unexpected joint speed clamping during execution.


\section{Conclusions}

We presented the implementation and performance evaluation of the ROS2 Fanuc CRX hardware interface.
We experimentally highlighted the strengths and limitations of the real-time control and communication framework based on the Fanuc Ethernet-IP module.
Despite the limited applicability to high-frequency closed-loop control, experiments show good results in setpoint reaching, path following, collision avoidance, and dynamic velocity modification.
Future works will explore other Fanuc control functions, such as the Dynamic Path Modification (DPM), to reduce communication latency.

\newpage

\clearpage



\section*{Supplementary material}

## Description of Supplementary Materials

| File / Resource | Description |
|---|---|
| **Core Folders** | |
| `TP_programs/` | Basic TP programs enabling communication and motion of the real Fanuc robot. Must be running before launching the hardware interface. |
| `crx_description/` | Contains URDF, XACRO, and mesh files for the robot model. Easily customizable for different cell layouts. |
| `crx_moveit_config/` | MoveIt2 configuration package for planning and execution. It should be adjusted for new scenes. |
| `fanuc_control/` | Core hardware interface logic implementation for robot communication and control. |
| `fanuc_eth_ip/` | Implements Ethernet/IP communication between PC and Fanuc controller. |
| `fanuc_rmi/` | RMI communication alternative (EXPERIMENTAL). Limited testing; included for advanced users. |
| `fanuc_srvs/` | ROS2 service examples for I/O control (e.g., Robotiq gripper). Requires additional robot-side logic. |
| `experiments/` | Python scripts to reproduce the experiments: `step_response.py` – step response `sinusoidal_traj.py` – sinusoidal trajectory `add_obstacle.py` – add obstacle in MoveIt2 scene |
| **Launch Parameters** | |
| `robot_type` | One of [crx5ia, ..., crx30ia]; model of the CRX robot. |
| `use_mock_hardware` | true/false – Enables virtual vs. real robot control. If true, it integrates a fake hardware interface with MoveIt2/RViz. |
| `controllers_file` | Path to the YAML configuration file for ROS2 controllers. |
| `robot_ip` | IP address of the physical robot (required if not using mock hardware). |
| `read_only` | true/false – Enables passive state monitoring without running a TP program. |
| `use_rmi` | true/false – EXPERIMENTAL; enables RMI instead of Ethernet/IP. |
| **Available Topics** | |
| `/cmd_j_pos` | Joint command input topic from `ros2_controllers`. Converts radians to degrees and accounts for J2/J3 coupling. |
| `/fb_j_pos` | Joint state feedback including positions and velocities (same conversion as above). |
| `/fb_c_pos` | Cartesian feedback from the Fanuc controller. Format: {x, y, z, Rx, Ry, Rz}; units: mm and degrees. |
| **System Requirements** | |
| Ubuntu 22.04 | Target operating system. |
| ROS 2 Humble | Target ROS version. |
| ROS packages | `ros-humble-ros2-control`, `ros-humble-moveit`. |
| EIPScanner | Library required for Ethernet/IP communication with the Fanuc controller. |
| ros2_fanuc_interface | Main ROS2 package containing robot interface code. |
| scaled_fjt_controller | Velocity-scaled controller used during experiments. |
| **Reproducing the Experiments** | |
| Launch robot | `ros2 launch fanuc_control robot_bringup.launch.py`<br>`    robot_ip:=your.robot.ip.address`<br>`    robot_type:=crx25ia`<br>`    controllers_file:=scaled_velocity_controller.yaml` |
| Step response | `python experiments/scripts/step_response.py` |
| Trajectory following | `python experiments/scripts/sinusoidal_traj.py` |
| Obstacle avoidance | `python experiments/scripts/add_obstacle.py`, then move robot in GUI |
| Velocity scaling | `ros2 topic pub /speed_ovr std_msgs/msg/Int16 "{data: vel_scale}"` where `vel_scale` ∈ [0,100] |



\vfill

\end{document}